\documentclass[letterpaper]{article} 
\usepackage{aaai2026}  
\usepackage{times}  
\usepackage{helvet}  
\usepackage{courier}  
\usepackage[hyphens]{url}  
\usepackage{graphicx} 
\usepackage{natbib}  
\usepackage{caption} 
\usepackage{algorithm}
\usepackage{algorithmic}
\usepackage{times}  
\usepackage{helvet}  
\usepackage{courier}  
\usepackage[hyphens]{url}  
\usepackage{graphicx} 
\usepackage{booktabs}
\usepackage{siunitx}
\usepackage{natbib}  
\usepackage{caption} 
\usepackage{algorithm}
\usepackage{algorithmic}
\usepackage{amsmath, amssymb}
\usepackage{multirow} 
\usepackage{newfloat}
\usepackage{amsmath}
\usepackage{graphicx}
\usepackage{caption} 
\usepackage{cuted}
\usepackage{listings}
\floatstyle{ruled}
\newfloat{listing}{tb}{lst}{}
\floatname{listing}{Listing}
\usepackage{newfloat}
\usepackage{listings}
\DeclareCaptionStyle{ruled}{labelfont=normalfont,labelsep=colon,strut=off} 
\floatstyle{ruled}
\newfloat{listing}{tb}{lst}{}
\floatname{listing}{Listing}
\title{EchoGen: Cycle-Consistent Learning for Unified Layout-Image Generation and Understanding} 

\author {
    Kai Zou\textsuperscript{\rm 1,6},
    Hongbo Liu\textsuperscript{\rm 2},
    Dian Zheng\textsuperscript{\rm 3},
    Jianxiong Gao\textsuperscript{\rm 4},
    Zhiwei Zhao\textsuperscript{\rm 5},
    Bin Liu\textsuperscript{\rm 1,6}\thanks{Corresponding authors}
}
\affiliations {
    \textsuperscript{\rm 1}School of Cyber Science and Technology, University of Science and Technology of China\\
    \textsuperscript{\rm 2}Tongji University\\
    \textsuperscript{\rm 3}Sun Yat-sen University\\
    \textsuperscript{\rm 4}Fudan University\\
    \textsuperscript{\rm 5}Hefei University of Technology\\
    \textsuperscript{\rm 6}Anhui Province Key Laboratory of Digital Security\\
    kzou@mail.ustc.edu.cn, 1952350@tongji.edu.cn, zhengd35@mail3.sysu.edu.cn,\\
    jxgao22@m.fudan.edu.cn, zhiweizhao@hfut.edu.cn, flowice@ustc.edu.cn
}

\usepackage{bibentry}

\begin{document}

\maketitle

\begin{abstract}
In this work, we present EchoGen, a unified framework for layout-to-image generation and image grounding, capable of generating images with both accurate layout and high fidelity to the text description.(\textit{e.g.,} spatial relationship), and grounding the image robustly at the same time. We believe that image grounding possesses strong text and layout understanding abilities, which can compensate for the corresponding limitations in layout-to-image generation. At the same time, images generated from layouts exhibit high diversity in content, thereby enhancing the robustness of image grounding. Jointly training both tasks within a unified model can promote performance improvements for each. However, we identify that this joint training paradigm encounters several optimization challenges and results in restricted performance. To address these issues, we propose progressive training strategies. First, the Parallel Multi-Task Pre-training (PMTP) stage equips the model with basic abilities for both tasks, leveraging shared tokens to accelerate training. Next, the Dual Joint Optimization (DJO) stage exploits task duality to sequentially integrate the two tasks, enabling unified optimization. Finally, the Cycle RL stage eliminates reliance on visual supervision by using consistency constraints as rewards, significantly enhancing the model’s unified capabilities via the GRPO strategy. Extensive experiments demonstrate state-of-the-art results on both layout-to-image generation and image grounding benchmarks, and reveal clear synergistic gains from optimizing the two tasks together.
\end{abstract}





\section{Introduction}

\begin{figure*}[h!t]
  \centering
   \includegraphics[width=.9\linewidth]{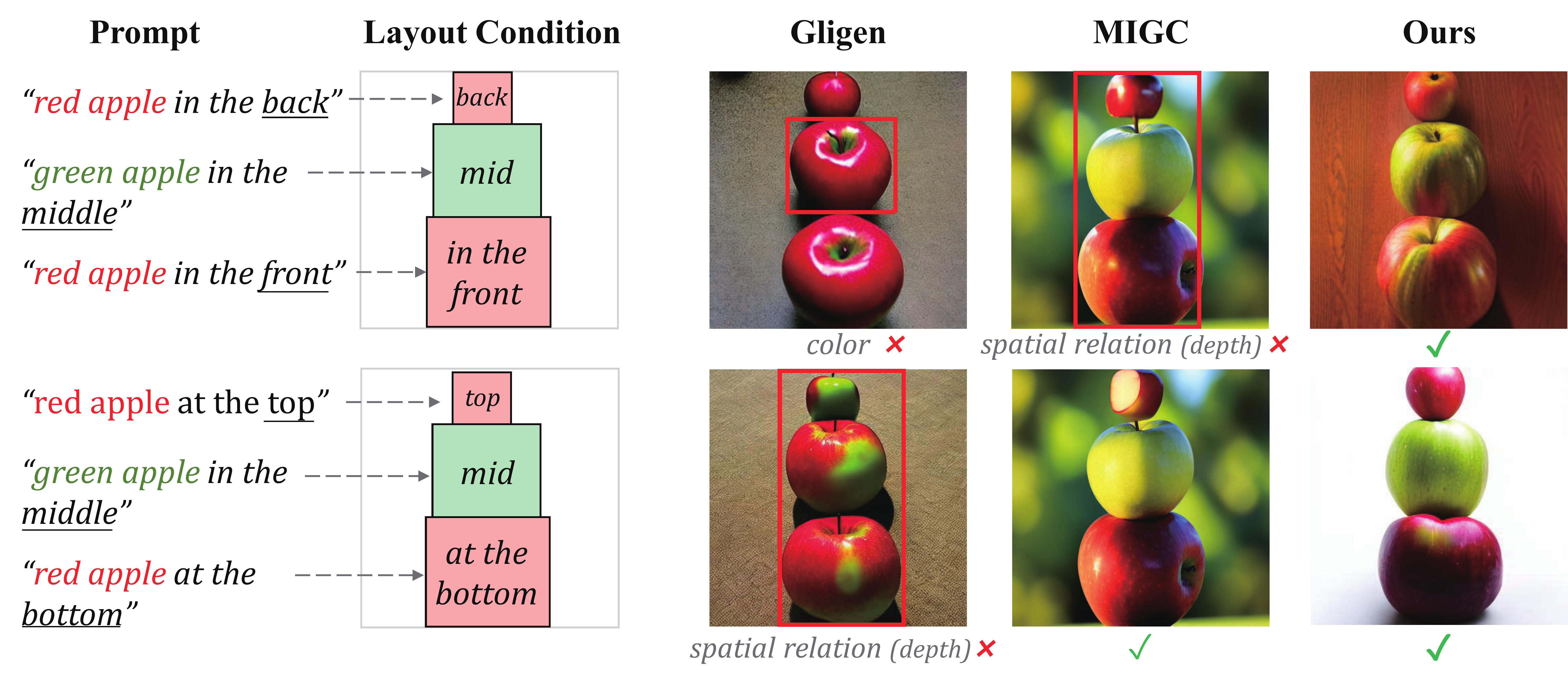}

   \caption{
    Our unified approach faithfully parses complex prompts—e.g., color attributes and relative spatial relations—and, combined with layout conditions, achieves more accurate semantic alignment. In contrast, specialized methods (e.g., \textit{Gligen} and \textit{MIGC}) often fail to handle such constraints. For instance, in the first row of \textit{MIGC} results, the model generates a vertical spatial relation (top–middle–bottom), whereas the prompt specifies a depth-based relation (front–middle–back).
    }

   \label{fig:tease}
\end{figure*}

Applications such as poster generation~\cite{chen2025postercraft}, interactive image editing~\cite{liu2025magicquill}, and visual question answering~\cite{antol2015vqa} demand both controllable image generation and accurate visual understanding. Two fundamental tasks in this domain are layout-to-image generation~\cite{zhou2024migc} and image grounding~\cite{grounding_dino}. The former aims to synthesize realistic images conditioned on grounding text and spatial layout, while the latter reverses this process by identifying and describing visual elements in an image, mapping them back to structured text and layout.

Currently, researchers mainly handle the two tasks independently~\cite{grounding_dino, kosmos_2}, lacking a unified architecture and training paradigm for joint optimization. However, they ignore the fact that 
joint training enables the model to naturally acquire favorable properties that are difficult to achieve in single-task settings. For example, as shown in Fig.~\ref{fig:tease}, we aim to generate an image of three apples placed on a table, where their pixel positions are specified by bounding boxes, and their \textit{spatial arrangement}—such as front-to-back or top-to-bottom—is described by the prompt.
Single-task approaches struggle to accurately distinguish and interpret such spatial relationships conveyed in the text.
In contrast, our unified framework benefits from the grounding task, which explicitly requires the model to understand spatial descriptions—such as identifying the nearest apple—thereby naturally enhancing alignment to both textual semantics and spatial layout specifications.

 In this work, we propose EchoGen, a unified framework for layout-to-image generation and image grounding, aiming at exhibiting the synergistic effect between two tasks. In practice, we found that directly jointly training the two tasks leads to limited performance; therefore, we propose progressive training strategies. First, we perform Parallel Multi-Task Pre-training (PMTP), accelerate the learning of base capabilities by share visual token representations, leveraging the inherent duality between generation and grounding tasks.  Next, we serialize the two tasks into a joint learning objective to achieve synergistic optimization and strengthen consistency along the layout–image–layout cycle mapping, which primes the model for the self‑supervised Reinforcement Learning (RL) stage.  Finally, by treating the grounding discrepancy as rewards, we perform self-supervised RL over the layout→image→layout loop (CycleRL) without requiring explicit supervision of the visual output, relying solely on text prompts with even random bounding boxes.

 Extensive experiments on MS-COCO, LayoutSAM, and Ref-L4 demonstrate the effectiveness of our approach on both layout-to-image generation and image grounding. 

 The key contributions of this paper are as follows:
\begin{itemize}
\item We propose a unified training framework for layout-to-image generation and image grounding, named EchoGen, enabling collaborative learning and efficient training with a duality-driven joint objective for the two tasks.
\item We achieve self-supervised RL over the layout$\to$image$\to$layout loop using cycle-consistent grounding discrepancy as rewards, eliminating the need of explicit visual supervision.
\item Experiments on MS-COCO, LayoutSAM, and Ref-L4 verify state-of-the-art performance and clear synergistic gains.
\end{itemize}

\section{Related Works}
\subsection{Layout Control in Image Generation}

Pure text guidance in image generation lacks precise spatial control, necessitating explicit layout conditions (e.g., bounding boxes, segmentation maps). \emph{Training-free} methods typically edit cross-attention~\shortcite{bartal2023multidiffusion} to control text-image interactions, or apply energy-based latent updates~\shortcite{xie2023boxdiff,xiao2023rb,chen2023trainingfree}; however, these approaches often struggle with complex layouts.  In contrast, \emph{training-based} approaches~\shortcite{hoe2024interactdiffusion,yang2022reco,Avrahami_2023,zheng2023layoutdiffusion} introduce explicit conditioning modules. For instance, \textit{GLIGEN}~\shortcite{li2023gligen} injects grounding tokens and spatial boxes; \textit{InstanceDiffusion}~\shortcite{wang2024instancediffusion} utilizes mask-guided conditioning; \textit{MIGC}~\shortcite{zhou2024migc} perform optimization in a divide-and-conquer way. Methods like \textit{IFAdapter}~\shortcite{wu2024ifadapter} and \textit{ELIGen}~\shortcite{zhang2025eligen} transfer control to more powerful backbones, balancing fidelity and controllability. Unified models (\textit{PlanGen}~\shortcite{he2025plangen}) integrate planning, generation, and grounding within an auto-regressive model, but they still optimize each task in isolation, resulting in suboptimal results. 
\subsection{Image Grounding}

Image grounding aims to localise textual references in images~\shortcite{grounded_language_image_pretraining,grounding_dino,kosmos_2}. Early open-vocabulary detectors such as GLIP~\shortcite{grounded_language_image_pretraining}, Grounding DINO~\shortcite{grounding_dino} and the real-time YOLO-World~\shortcite{cheng2024yolo} combine detection backbones with text features to support zero-shot phrase localisation. More recently, general-purpose vision–language models like CogVLM~\shortcite{cog_vlm} and Qwen2.5-VL~\shortcite{bai2025qwen2} are trained on large‐scale visual–text corpora with bounding-box supervision and achieve strong zero-shot results on standard grounding benchmarks\shortcite{ref_coco_+, chen2025revisiting}. 
We further incorporate image grounding into a unified framework for multimodal understanding and generation, enabling joint optimization with layout-to-image synthesis.
\subsection{Unified Multimodal Models}
Recently, there is growing interest in unified multimodal models that can both understand and generate visual content within one architecture~\shortcite{zhang2025unified_survey,wu2025janus,deng2025emerging,zou2025uni}. 
Some approaches~\shortcite{zhou2024transfusion} combines an autoregressive language modeling objective with an image diffusion process to train one Transformer capable of producing either text or images from a given prompt. Janus ~\shortcite{wu2025janus} addresses the different granularity needs between understanding and generation by decoupling its visual encoder into separate pathways, while still using a shared Transformer backbone. Building on this, Janus-Pro~\shortcite{chen2025janus} yields further gains in multimodal understanding and generation. Most recently, BAGEL~\shortcite{deng2025emerging} demonstrated that a decoder-only model pre-trained on trillions of interleaved text-image-video tokens can exhibit emergent multimodal capabilities.

\begin{figure*}[h!t]
  \centering
   \includegraphics[width=1.0\linewidth]{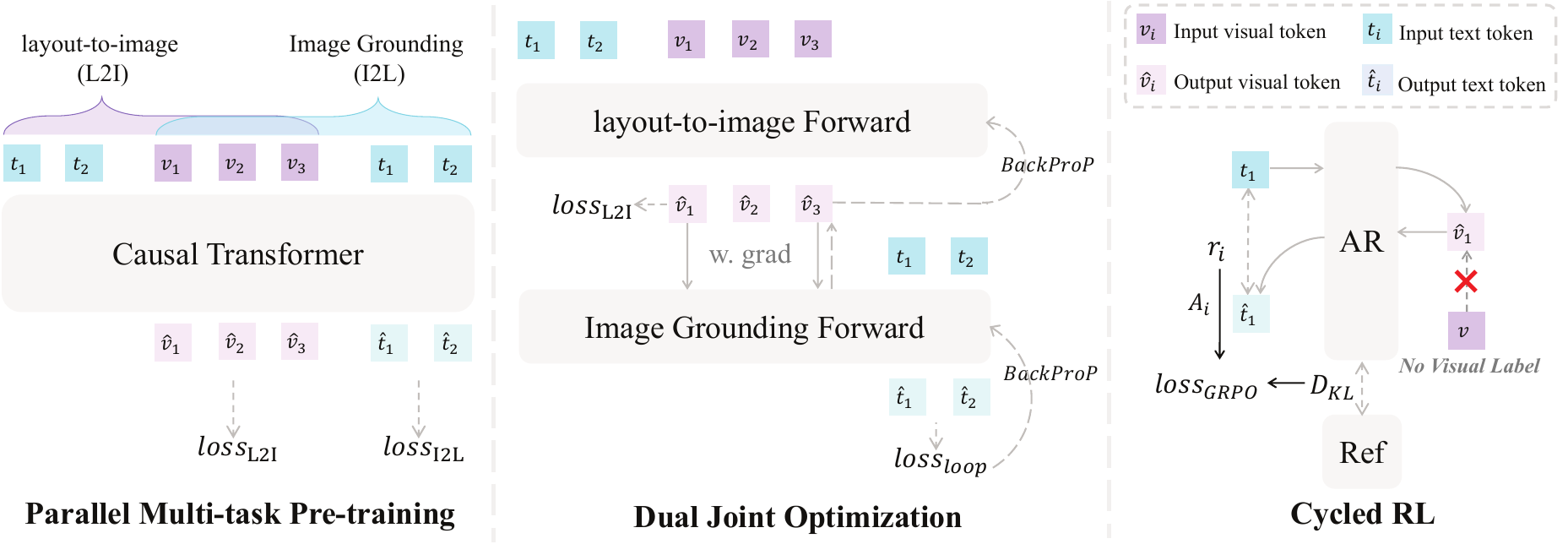}

   \caption{Overview of \emph{EchoGen}.
    \textbf{Left—Parallel Multi‑task Pre‑training:} a unified autoregressive transformer is trained on layout‑to‑image and image grounding tasks in parallel, yielding fast acquisition of base capabilities. 
    \textbf{Middle—Dual Joint Optimization:} the image tokens generated by the generation forward pass are directly reused as the input to grounding, forming a single joint objective, which strengthens layout$\to$image$\to$layout cycle consistency. 
    \textbf{Right—Cycled RL:} we execute the layout$\to$image$\to$layout loop and treat the discrepancy between input and recovered layouts as a continuous reward, enabling self-supervised Reinforcement Learning without explicit supervision of intermediate visual outputs.}

   \label{fig:pipeline}
\end{figure*}

\section{Method}

\subsection{Overview}
To enable fully self-supervised reinforcement learning in the final stage, the model must first acquire reliable layout-to-image synthesis and image grounding capabilities and exhibit sufficient consistency along the layout$\to$image$\to$layout (L--I--L) loop. Accordingly, we organize training into three stages, as shown in Fig.~\ref{fig:pipeline}. \emph{First}, a Parallel Multi-task Pretraining phase establishes base abilities for both tasks while sharing image-token pathways to accelerate training. \emph{Second}, a Dual Joint Optimization phase nests generation and grounding by feeding the synthesized image directly into the grounding branch, forming a single joint objective that strengthens L--I--L cycle consistency. \emph{Finally}, a Cycled Reinforcement Learning stage executes the L--I--L loop and treats the spatial discrepancy between input and recovered layouts as a continuous reward, enabling self-supervised optimization without explicit supervision of intermediate visual outputs.


\subsection{Parallel Multi-task Pre-training}
\begin{figure}[h!t]
  \centering
   \includegraphics[width=.8\linewidth]{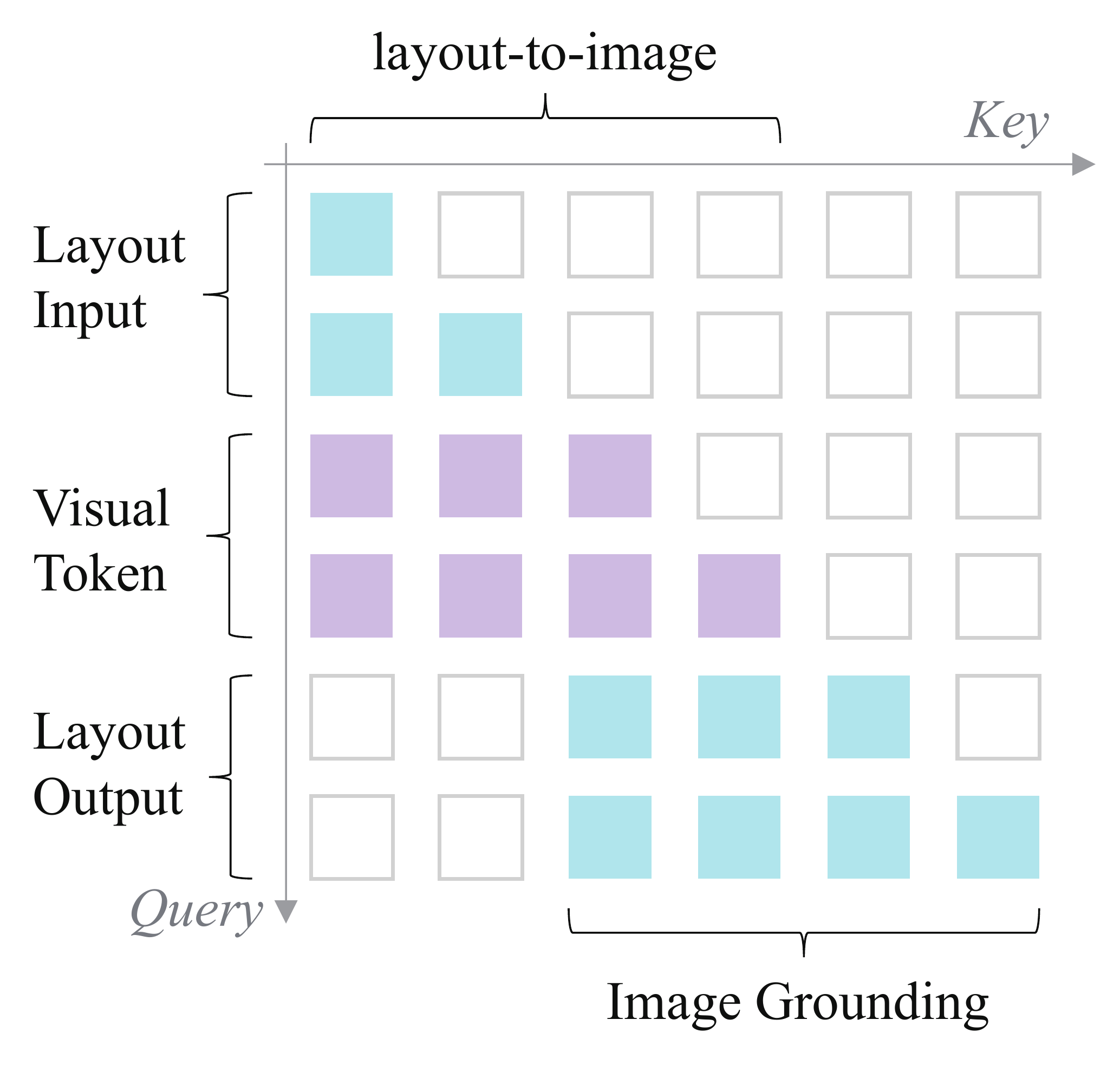}

   \caption{
    Attention Mask of parallel multi-task pre-training.
    }

   \label{fig:mask}
\end{figure}

Pretrained unified models often lack native layout-to-image synthesis capabilities or exhibit insufficient grounding proficiency. To prepare for subsequent stages, we introduce a multi-task learning phase that jointly acquires both layout-to-image and grounding abilities. 

In both tasks, the proportion of visual tokens is overwhelmingly large—typically on the order of hundreds or even thousands—whereas the number of text tokens is usually only in the tens. Moreover, a single image can simultaneously serve as the output of layout-to-image synthesis and the input of image grounding. Leveraging this duality, we concatenate the inputs of two tasks at the token level, so that a large subset of visual tokens is shared across tasks, thereby enabling efficient parallel learning. Concretely, given a grounding–image pair \((X_{\mathrm{g}}, X_{\mathrm{i}})\), we form the input sequence
\(
\bigl[X_{\mathrm{g}},\;X_{\mathrm{i}},\;X_{\mathrm{g}}\bigr]
\)
and feed it into an autoregressive Transformer, which produces the predicted sequence
\(
\bigl[Y_{\mathrm{g}},\;Y_{\mathrm{i}},\;Y_{\mathrm{g}}\bigr].
\)
As shown in Fig.~\ref{fig:mask}, a task-aware attention-mask is designed to prevent cross-task information leakage. We then optimize the summed next-token prediction losses:
\begin{equation}
\mathcal{L}_{\mathrm{pretrain}}
=\;\mathrm{CE}\bigl(X_{\mathrm{i}},\,Y_{\mathrm{i}}\bigr)
\;+\;\mathrm{CE}\bigl(X_{\mathrm{g}},\,Y_{\mathrm{g}}\bigr),
\end{equation}
where CE denotes the cross-entropy loss function. This design efficiently enables the unified model with both layout-to-image and grounding capabilities.

\subsection{Dual Joint Optimization}

At this stage, we exploit the input–output duality between layout-to-image generation (L2I) and image grounding (I2L) by \emph{nesting} them into a layout$\to$image$\to$layout (L--I--L) loop, yielding a joint learning objective for collaborative optimization that strengthens cycle-mapping consistency and prepares the model for the subsequent self-supervised RL stage.

Let $\mathcal{L}$ and $\mathcal{I}$ denote the layout- and image-token spaces, respectively. A unified model $p_\theta(\cdot\,|\,\cdot)$ with shared parameters $\theta$ implements $f_\theta:\mathcal{L}\!\to\!\mathcal{I}$ and $g_\theta:\mathcal{I}\!\to\!\mathcal{L}$. On a dataset $\mathcal{D}=\{(\ell,i)\}$, we define the supervised (token-level) cross-entropy losses
\[ \mathcal{L}_{\text{I2L}}(\theta)=\mathbb{E}_{(\ell,i)\sim\mathcal{D}}\!\left[-\log p_\theta(\ell\,|\,i)\right], \] 
\[ \mathcal{L}_{\text{L2I}}(\theta)=\mathbb{E}_{(\ell,i)\sim\mathcal{D}}\!\left[-\log p_\theta(i\,|\,\ell)\right] \]
and the loop loss
\[
\mathcal{L}_{\text{loop}}(\theta)
=\mathbb{E}_{\ell}\!\Big[-\log p_\theta\!\big(\ell\,\big|\,\hat{\imath}\big)\Big],
\quad \text{where}\quad
\hat{\imath}\sim p_\theta(\cdot\,|\,\ell).
\]
\textit{Remark.} $\mathcal{L}_{\text{loop}}$ is also a (sequence-wise) cross-entropy objective that measures the negative log-likelihood of reconstructing $\ell$ from the sampled intermediate image tokens $\hat{\imath}$.

Notably, any global minimizer of $\mathcal{L}_{\text{L2I}}+\mathcal{L}_{\text{I2L}}$ on a noise-free $\mathcal{D}$ is also a global minimizer of $\mathcal{L}_{\text{loop}}$. Specifically, at a global optimum $\theta^\star$ we have $p_{\theta^\star}(i\,|\,\ell)=\delta_{F(\ell)}$ and $p_{\theta^\star}(\ell\,|\,i)=\delta_{G(i)}$, hence sampling $\hat{\imath}\!\sim\!p_{\theta^\star}(\cdot|\ell)$ yields $\hat{\imath}=F(\ell)$ a.s., and therefore $p_{\theta^\star}(\ell\,|\,\hat{\imath})=1$, making $\mathcal{L}_{\text{loop}}(\theta^\star)=0$.

However, minimizing $\mathcal{L}_{\text{loop}}$ alone does not preclude \emph{shortcuts} that bypass faithful image synthesis (e.g., mapping $\ell$ to arbitrary ``code'' tokens that decode back to $\ell$). We thus regularize the loop objective with the explicit visual supervision of L2I by the following joint loss:
\begin{equation}
\label{eq:joint}
\mathcal{J}_{\text{joint}}(\theta)
=\mathcal{L}_{\text{L2I}}(\theta)+\lambda\,\mathcal{L}_{\text{loop}}(\theta),\qquad \lambda>0.
\end{equation}
Both terms in $\mathcal{J}_{\text{joint}}$ are cross-entropy losses: $\mathcal{L}_{\text{L2I}}$ enforces likelihood of visually correct image tokens given the layout, while $\lambda\,\mathcal{L}_{\text{loop}}$ enforces likelihood of reconstructing the original layout after passing through the sampled image-token path. Compared with the standalone grounding loss, the \emph{nested} joint objective encourages ``generate then return'' consistency ($g_\theta\!\circ f_\theta\!\approx\!\mathrm{Id}$) and backpropagates through the same image-token pathway as L2I, yielding better gradient alignment and smoother updates that foster synergy between the two tasks.

Because sampling intermediate image tokens is non-differentiable in an autoregressive transformer, we adopt a Gumbel--Softmax approximation to maintain gradient connectivity along the L--I--L loop. At decoding step $t$ with logits $\mathbf{z}_t$, we draw Gumbel noise $\boldsymbol{\gamma}_t\!\sim\!\mathrm{Gumbel}(0,1)$ and form the differentiable sample
\[
\tilde{\mathbf{y}}_t=\mathrm{softmax}\!\left((\mathbf{z}_t+\boldsymbol{\gamma}_t)/\tau_t\right),
\]
using the straight-through estimator in the forward pass, while backpropagating through $\tilde{\mathbf{y}}_t$.
To balance exploration and discreteness, the temperature parameter follows an annealing schedule over training:
\[
\tau_t \equiv \tau(k)=\max\!\big(\tau_{\min},\,\tau_0\,\alpha^{\,k}\big),
\quad 0<\alpha<1,
\]
where $k$ indexes the update steps, $\tau_0$ is the initial temperature, and $\tau_{\min}$ prevents premature collapse. This schedule gradually sharpens $\tilde{\mathbf{y}}_t$ toward one-hot selections, preserving stable gradients early and near-discrete decoding later. Combined with \eqref{eq:joint}, this yields end-to-end differentiable, collaboratively optimized layout-to-image generation and image grounding.

\begin{table*}[ht]
\centering
\sisetup{detect-weight=true,detect-family=true}
\label{tab:coco-layoutsam}
\setlength{\tabcolsep}{6pt}
\renewcommand{\arraystretch}{1.15}
\begin{tabular}{l *{9}{S[table-format=2.2]}}
\toprule[0.12em]
\multicolumn{1}{l}{\multirow{2}{*}{\textbf{Method}}} & \multicolumn{5}{c}{\textbf{MS COCO}} & \multicolumn{4}{c}{\textbf{LayoutSAM-Eval}} \\
\cmidrule(lr){2-6} \cmidrule(l){7-10}
 &
{\(\mathrm{AP}\)$\uparrow$} & {\(\mathrm{AP}_{50}\)$\uparrow$} & {\(\mathrm{AP}_{75}\)$\uparrow$} &
{CLIP$\uparrow$} & {FID$\downarrow$} &
{Spatial$\uparrow$} & {Color$\uparrow$} & {Texture$\uparrow$} & {Shape$\uparrow$} \\
\midrule[0.1em]
\multicolumn{10}{c}{\textbf{\textit{Gen.\ only}}} \\
\midrule[0.1em]
GLIGEN             & 30.99 & 41.44 & 36.89 & 24.81 & 27.93 & 77.53 & 49.41 & 55.29 & 52.72 \\
MIGC               & 46.16 & 58.29 & 50.21 & 25.04 & 25.35 & 85.66 & 66.97 & 71.24 & 69.06 \\
InstanceDiffusion  & 49.97 & 61.33 & 52.65 & 25.15 & 25.00 & 87.99 & 69.16 & 72.78 & 71.08 \\
IFAdapter          & 43.31 & 56.29 & 52.76 & 26.39 & 23.45 & 84.32 & 68.32 & 71.37 & 68.37 \\
EliGen             & 45.83 & 59.26 & 49.66 & \textbf{27.44} & 21.43 & 85.37 & 72.35 & 75.42 & 74.59 \\
\midrule[0.1em]
\multicolumn{10}{c}{\textbf{\textit{Und.\ and Gen.}}} \\
\midrule[0.1em]
PlanGen            & 51.39 & 64.70 & 61.10 & 25.33 & 20.44 & 92.21 & 82.69 & 86.53 & 85.36 \\
\textbf{EchoGen (Ours)}      & \textbf{54.61} & \textbf{68.85} & \textbf{65.01} & 25.18 & \textbf{20.12} & \textbf{96.32} & \textbf{84.97} & \textbf{89.02} & \textbf{87.18} \\
\bottomrule[0.1em]
\end{tabular}
\caption{{Quantitative results on MS COCO and LayoutSAM-Eval benchmark.} Method marked as \textit{Gen.\ only} perform layout-to-image generation only, while \textit{Und.\ and Gen.} denotes unified layout-to-image generation and grounding models.}
\end{table*}

\subsection{Cycled Reinforcement Learning}
At this stage, the model has acquired reliable layout-to-image (L2I) and image-to-layout grounding (I2L) capabilities and exhibits sufficient layout$\to$image$\to$layout (L--I--L) cycle consistency. We therefore perform \emph{self-supervised} RL by using the L--I--L layout discrepancy as a continuous reward, without supervising the intermediate visual output. This is justified because the preceding stages ensure that RL samples yield meaningful visual signals, preventing degradation of the generation ability.

Concretely, let a grounding sequence be $X_{\mathrm{g}}=\{(x_{e}^{k},x_{b}^{k})\}_{k=1}^{K}$, where $x_{e}^{k}$ is the referring expression and $x_{b}^{k}$ the ground-truth box of the $k$-th instance. We draw an image sample $\hat{i}\!\sim\!p_{\theta}(\cdot\mid X_{\mathrm{g}})$ via L2I, and then ground each expression on $\hat{i}$ via I2L to obtain predicted boxes $\{\hat{y}_{b}^{k}\}_{k=1}^{K}$. Let $d(\cdot,\cdot)$ be a differentiable box discrepancy (e.g., $1{-}\mathrm{IoU}$, $\ell_{1}$), and define the per-sample aggregate discrepancy as a continuous reward
\begin{equation}
r_{bbox}=\frac{1}{K}\sum_{k=1}^{K} d(\hat{y}_{b}^{k},x_{b}^{k}).
\end{equation}

We apply GRPO~\cite{shao2024deepseekmath} over groups $\{r_{bbox}^i\}_{i=1}^{G}$, using the group-relative advantage $A_{bbox}^i=r_{bbox}^i-\frac{1}{G}\sum_{j=1}^{G} r^j$. Given an input query $q$, GRPO samples a set of candidate outputs $\{o^1, o^2, \dots, o^G\}$ using the current policy $\pi_{\text{old}}$. The new policy $\pi_\theta$ is optimized to maximize the following objective:

\begin{align}
\mathcal{J}(\theta) =\ 
& \mathbb{E}_{q \sim \mathcal{D},\, \{o^i\}_{i=1}^{G} \sim \pi_{\text{old}}(\cdot|q)} \Bigg[ \frac{1}{G} \sum_{i=1}^{G} \Big( \notag \\
& \min \Big( r_{bbox}^i A_{bbox}^i,\ \text{clip}(r_{bbox}^i,\ 1{-}\epsilon,\ 1{+}\epsilon) A_{bbox}^i \Big) \notag \\
&\quad - \beta\, \mathbb{D}_{\text{KL}}(\pi_\theta \,\|\, \pi_{\text{ref}}) \Big) \Bigg] \label{eq:grpo}
\end{align}

where $r_{bbox}^i = \frac{\pi_\theta(o^i|q)}{\pi_{\text{old}}(o^i|q)}$ is the importance weight, $A_{bbox}^i$ denotes the group-relative advantage, and $\beta$ is a weighting factor on the regularization term. Together with the KL regularizer, the policy is optimized by the GRPO objective, yielding self-supervised, looped reinforcement learning driven solely by cycle-consistent layout rewards.

\section{Experiments}

\subsection{Implementation Details}
The experiments are conducted with a pretrained \emph{Janus-Pro 1.5B}~\cite{chen2025janus} backbone on NVIDIA A100--80GB GPUs. 
Across all three stages, we use the AdamW optimizer with a fixed learning rate of 
$5\times 10^{-5}$; the mini‑batch size is 32 for Stages~1 and~2. 
In the reinforcement learning stage, we adopt Group Relative Policy Optimization (GRPO) with 
a group size of 8.
To ensure gradient propagation between the two tasks during Stage~2 (Dual Joint Optimization), 
we use the VQ‑VAE token sequence as the visual representation for image grounding. 
To preserve the model’s pretrained visual understanding, we randomly retain ViT‑encoded 
tokens with probability $50\%$, and we keep both the VQ‑VAE encoder and the ViT encoder 
frozen throughout all stages.

\begin{table}[t]
\centering

\begin{tabular}{lcc}
\toprule[0.12em]
\textbf{Stage} & \textbf{Samples} & \textbf{Steps} \\
\midrule[0.1em]
Parallel Multi‑task Pre‑training & 4M    & 125K  \\
Dual Joint Optimization          & 2M    & 60K  \\
Cycled RL                & 50K  & 50K \\
\bottomrule[0.1em]
\end{tabular}

\caption{Training recipe.}
\label{tab:train-schedule}
\end{table}

\noindent\textbf{Training data.} Starting from GRIT‑20M, we filter bounding boxes by confidence $\geq 0.8$ and discard images
containing fewer than 2 or more than 10 boxes. We then sample approximately 6M 
layout–image pairs for training. The number of samples and optimization steps per stage are 
summarized in Tab.~\ref{tab:train-schedule}.

\begin{figure*}[h!t]
  \centering
   \includegraphics[width=1.0\linewidth]{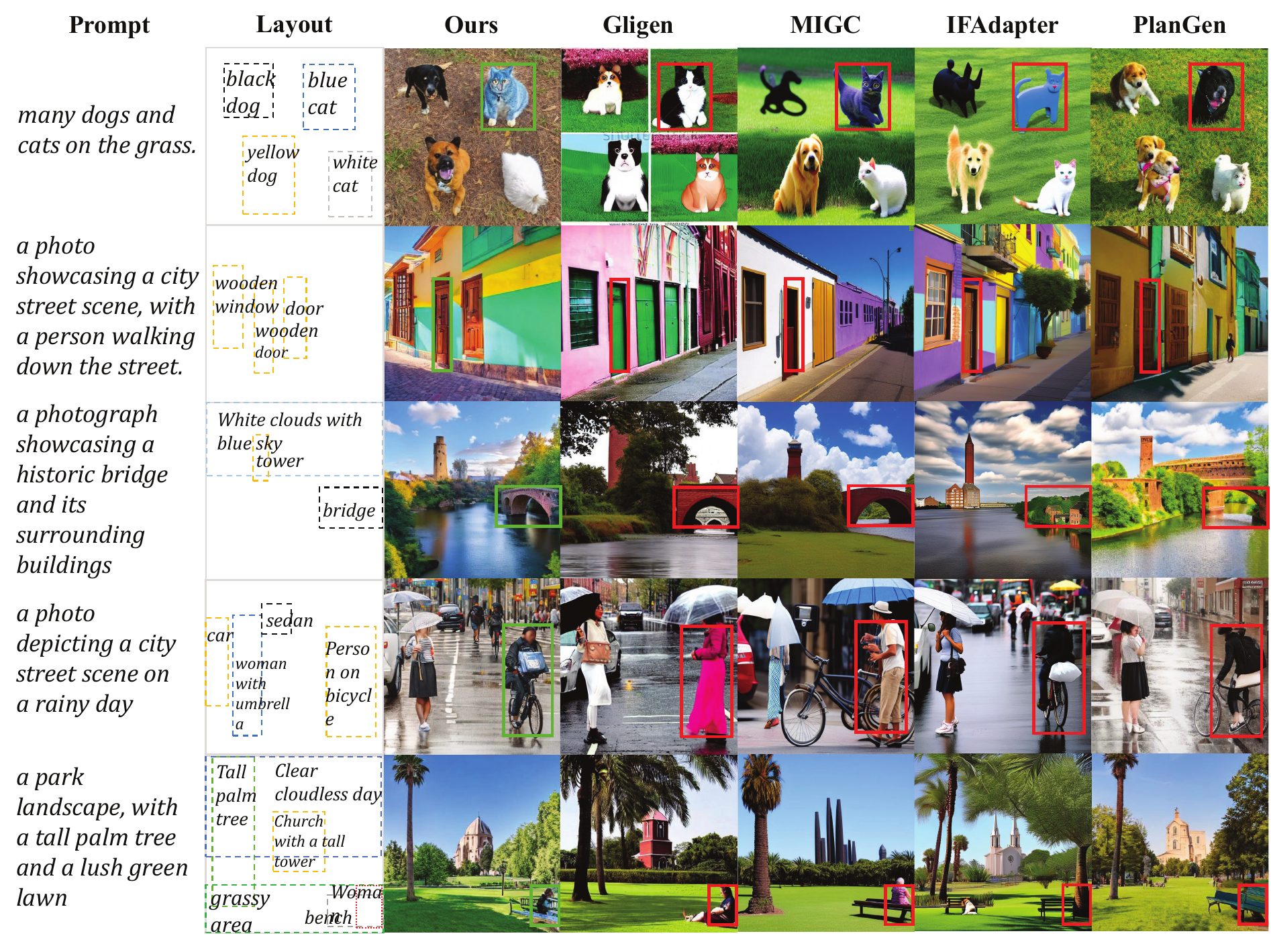}

   \caption{
    Quantitative comparison on layout-to-image generation. The first column lists the global text prompts; the second column visualizes the layout conditions, rendered as \emph{(referring-expression, bounding-box)} pairs. We highlight a subset of negatives and positives with red and green boxes; note that negatives may correspond to \emph{incorrect spatial locations or attributes (e.g., color)}.
    }

   \label{fig:quality}
\end{figure*}

\subsection{Experiments Setup}
\noindent\textbf{layout-to-image Evaluation.}
We evaluate our layout-to-image synthesis on \emph{MS-COCO}~\cite{lin2015microsoft} and \emph{LayoutSAM-Eval}~\cite{zhang2024creatilayout}.
For MS-COCO, we employ \emph{Grounding-SAM}~\cite{kirillov2023segment,ren2024grounded} to detect objects in the generated images and obtain bounding boxes; we set box threshold to 0.3 and text threshold to 0.25 without non-maximum suppression (NMS); standard COCO metrics are then computed for boxes, including $\text{AP}$, $\text{AP}_{50}$, and $\text{AP}_{75}$.
In addition, we report the \emph{CLIP} score~\cite{radford2021learning} to assess global text--image alignment, and the Fr\'echet Inception Distance (FID)~\cite{heusel2018gans} to quantify distributional image quality. For \emph{LayoutSAM-Eval}, we adopt \emph{MiniCPM-V-2\_6}~\cite{yao2024minicpm} as the evaluation vision--language model and report four dimensions---\textit{Spatial}, \textit{Color}, \textit{Texture}, and \textit{Shape}---which probe instance-level controllability of the layout-to-image generattion along spatial arrangement and appearance factors.

\noindent\textbf{Image Grounding Evaluation.} 
We assess grounding on the \emph{Ref-L4} benchmark~\shortcite{chen2025revisiting}, tailored to modern referring expression comprehension (REC) with long, information-dense queries and broad visual diversity. Following the official protocol, we report $\mathrm{Acc}_{0.5}$, $\mathrm{Acc}_{0.75}$, $\mathrm{Acc}_{0.9}$, and $\mathrm{mAcc}$. 

\noindent\textbf{Baselines.}
For \emph{layout-to-image} synthesis, we consider six representative methods  based on diffusion backbones or auto-regressive transformers. 
\emph{GLIGEN}~\cite{li2023gligen}, \emph{MIGC}~\cite{zhou2024migc}, and \emph{InstanceDiffusion}~\cite{wang2024instancediffusion} are all built upon \texttt{Stable~Diffusion~v1.5}~\cite{rombach2022highresolution}, augmenting the backbone with task-specific modules to ingest layout signals; 
\emph{IFAdapter}~\cite{wu2024ifadapter} is based on \texttt{Stable~Diffusion~XL}~\cite{podell2023sdxl} with an adapter that explicitly learns layout control; 
\emph{EliGen}~\cite{zhang2025eligen} follows a \emph{DiT}~\shortcite{peebles2023scalable} architecture and fine-tunes \texttt{FLUX-dev}, targeting precise \emph{instance-level} controllability; 
and \emph{PlanGen}~\cite{he2025plangen} serves as a unified generation--understanding model based on auto-regressive transformers that supports both layout conditioning and image understanding. 
For \emph{image grounding}, we evaluate three large vision--language models—\emph{KOSMOS-2}\shortcite{kosmos_2}, Qwen-VL-Chat~\shortcite{bai2023qwen}, and \emph{CogVLM-grounding}~\shortcite{wang2024cogvlm}.

\subsection{Layout-to-image Generation}

\noindent\textbf{Quantitative results.}
On \emph{MS-COCO}, our approach surpasses both diffusion-based and unified-model baselines in layout-control accuracy. 
Relative to the previous best, we obtain absolute gains of 3.22 in $\mathrm{AP}$, 4.15 in $\mathrm{AP}_{50}$, and 3.92 in $\mathrm{AP}_{75}$. 
Our method also achieves higher CLIP scores than \texttt{SD1.5}-based methods and \emph{PlanGen}, indicating stronger prompt faithfulness. 
The remaining CLIP gap to \emph{IFAdapter} and \emph{EliGen} is plausibly explained by their stronger base models---\texttt{SDXL} and \texttt{FLUX-dev}---with substantially larger parameter counts (3.5B and 12B). 
In terms of image quality, our method attains the best Fr\'echet Inception Distance (FID). 
Notably, \emph{EliGen}, \emph{PlanGen}, and our model all maintain a clear FID advantage over the other baselines; we hypothesize this stems from their Transformer-based design that treats the layout condition as contextual tokens, in contrast to approaches that add extra modules to enforce explicit spatial constraints on latent-space, which can introduce artifacts and degrade perceptual quality.
On \emph{LayoutSAM-Eval}, our approach establishes a new state of the art, improving the \textit{Spatial}, \textit{Color}, \textit{Texture}, and \textit{Shape} dimensions by 4.11, 2.28, 2.49, and 1.82, respectively, with the largest gain observed for spatial control.

\noindent\textbf{Qualitative results.}
Figure~\ref{fig:quality} presents visual comparisons under diverse layout conditions.
For \emph{complex} specifications (e.g., Row~1 with multiple cats and dogs of distinct colors), \emph{GLIGEN}, \emph{MIGC}, and \emph{IFAdapter} tend to yield lower perceptual fidelity with overly smoothed, cartoon‑like textures, whereas our method maintains consistently high image quality.
Across all settings, our approach achieves precise \emph{spatial control} (placement and scale) and accurate \emph{attribute control} (e.g., color), faithfully adhering to the specified layouts.

\begin{table}[ht]
\centering
\setlength{\tabcolsep}{6pt}
\renewcommand{\arraystretch}{1.15}
\begin{tabular}{lcccc}
\toprule[0.12em]
\multirow{2}{*}{\textbf{Method}} & \multicolumn{4}{c}{\textbf{Ref-L4}} \\
& $\mathrm{Acc}_{0.5}$ & $\mathrm{Acc}_{0.75}$ & $\mathrm{Acc}_{0.9}$ & $\mathrm{mAcc}$ \\
\midrule[0.1em]
\multicolumn{5}{c}{\textbf{\textit{Und.\ only}}} \\
\midrule[0.1em]
KOSMOS-2             & 48.53 & 38.34 & 17.54 & 34.72 \\
Qwen-VL-Chat         & 73.80 & 58.05 & 37.16 & 55.94 \\
CogVLM-g.     & 81.70 & 70.77 & \textbf{48.35} & 66.09 \\
\midrule[0.1em]
\multicolumn{5}{c}{\textbf{\textit{Und.\ and Gen.}}} \\
\midrule[0.1em]
\textbf{Ours}        & \textbf{83.20} & \textbf{75.42} & 47.32 & \textbf{68.46} \\
\bottomrule[0.1em]
\end{tabular}
\caption{Quantitative results on Ref-L4. Method marked as \textit{Und.\ only} perform image grounding only, while \textit{Und.\ and Gen.} denotes unified layout-to-image generation and grounding models.}
\label{tab:ref-l4}
\end{table}

\subsection{Image Grounding}
Tab.~\ref{tab:ref-l4} summarizes the quantitative  comparison on Ref-L4.
Despite a  smaller parameter budget,  our method EchoGen achieves the best overall performance.
Compared to the previous state of the art, we achieve absolute gains of 1.50 in $\mathrm{Acc}_{0.5}$, 4.65 in $\mathrm{Acc}_{0.75}$, and 2.37 in $\mathrm{mAcc}$. These results highlight the synergistic effect of our unified architecture and training strategy, enabling faster performance gains with reduced training resources.

\subsection{Ablation Study}

\noindent\textbf{Ablation for training stage.}
We compare different stage configurations on MS‑COCO (Tab.~\ref{tab:stage_ablation}).
Adding \emph{Stage 2} on top of \emph{Stage 1} (\#1$\rightarrow$\#2) yields a sizeable gain of 5.12 AP (from 47.26 to 52.38), and introducing \emph{Stage 3} thereafter (\#2$\rightarrow$\#4) provides a further 2.23 AP improvement, totaling 7.35 over \#1.
By contrast, skipping Stage 2 and directly applying Stage 3 (\#1$\rightarrow$\#3) produces only a marginal AP increase (0.75), underscoring the necessity of Stage 2 for strengthening layout–image–layout consistency.
Regarding image quality, FID varies modestly across stages; notably, Stage 3 consistently reduces FID, which we hypothesize arises from the additional diversity introduced by stochastic sampling.

\begin{table}[t]
\centering
\setlength{\tabcolsep}{8pt}
\begin{tabular}{cccccc}
\toprule[0.12em]
 & \textbf{Stage 1} & \textbf{Stage 2} & \textbf{Stage 3} & \textbf{AP}$\uparrow$ & \textbf{FID}$\downarrow$ \\
\midrule[0.1em]
\#1 & \checkmark &               &               & 47.26 & 21.98 \\
\#2 & \checkmark & \checkmark     &               & 52.38 & 22.23 \\
\#3 & \checkmark &               & \checkmark     & 48.01 & 20.32 \\
\#4 & \checkmark & \checkmark     & \checkmark     & \textbf{54.61} & \textbf{20.12} \\
\bottomrule[0.1em]
\end{tabular}
\caption{Ablation over training stages.}
\label{tab:stage_ablation}
\end{table}

\begin{figure}[t]
  \centering
   \includegraphics[width=1.0\linewidth]{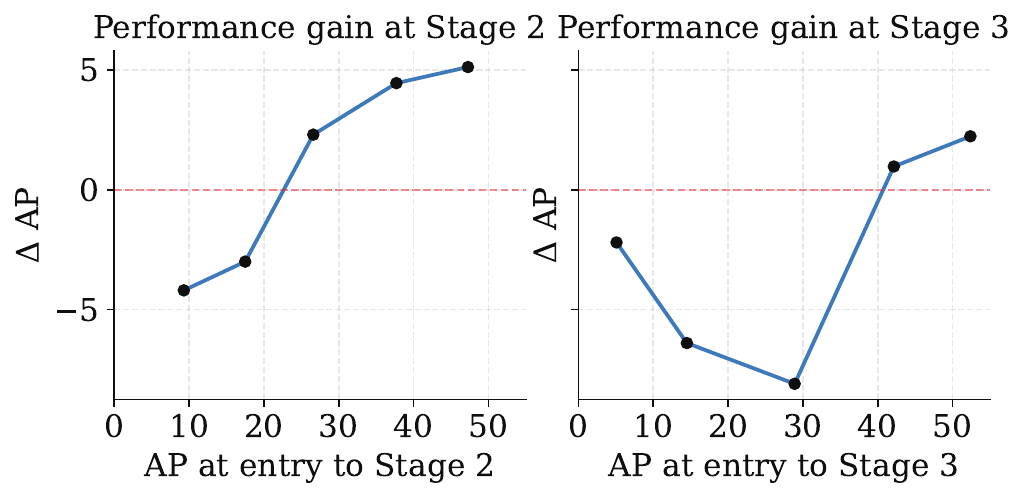}

   \caption{
    Ablation on stage transition points. Left: gains when switching from Stage~1$\!\rightarrow$Stage~2; Right: gains when switching from Stage~2$\!\rightarrow$Stage~3.
The x\mbox{-}axis records the $\mathrm{AP}$ at entry to the current stage (determined by the preceding stage’s training budget), while the y\mbox{-}axis reports the $\mathrm{AP}$ improvement achieved by that stage.
    }

   \label{fig:stage}
\end{figure}

\noindent\textbf{Ablation on stage transition points.}
To examine how the entry point into later stages affects performance, we vary the training budget of each stage on MS‑COCO and measure downstream outcomes under a fixed budget for the subsequent stage.
Fig.~\ref{fig:stage}\,(left) plots the \emph{post‑Stage\,2} AP against the AP achieved \emph{right before} switching from Stage\,1 to Stage\,2.
When the base capability is low (e.g., AP $<20$), Stage\,2 yields \emph{negative transfer}, likely because poor layout‑to‑image generations induce misaligned supervision on the image grounding branch, leading to incorrect gradient signals.
As the base AP improves, Stage\,2 confers increasingly larger gains, indicating that collaborative training is most effective once the generator reaches a minimally reliable quality. Fig.~\ref{fig:stage}\,(right) shows an analogous study for the Stage\,2$\rightarrow$Stage\,3 transition.
Since Stage\,3 (RL) does not provide explicit token‑level supervision on visual tokens, models entering with weaker base accuracy (AP $<40$) suffer pronounced performance drops, consistent with the difficulty of sampling sufficiently accurate generations to obtain informative rewards and gradients.
Note that AP has a practical lower bound of roughly $\sim3.0$. Consequently, when the pre‑switch AP is extremely low, the absolute drop is capped by this floor, so the measured degradation appears smaller; as the baseline AP increases toward $\sim30$, the potential absolute decrease correspondingly grows.

\begin{table}[t]
\centering
\setlength{\tabcolsep}{8pt}
\begin{tabular}{lccc}
\toprule[0.12em]
\textbf{Dataset} & \textbf{AP} & $\mathbf{AP}_{\mathbf{50}}$ & $\mathbf{AP}_{\mathbf{75}}$ \\
\midrule[0.1em]
From real dataset   & 54.61 & 68.85 & 65.01 \\
Random              & 53.96 & 68.02 & 64.32 \\
Generated by GPT\textendash4o & 54.51 & 68.94 & 64.59 \\
\bottomrule[0.1em]
\end{tabular}
\caption{Ablation on Stage\,3 (RL) training set.}
\label{tab:text_source_ablation}
\end{table}

\noindent\textbf{Ablation on Stage\,3 training set.}
Stage\,3 (RL) can exploit \emph{layout text alone} as a self‑supervisory signal, obviating the need for paired images.
To examine whether the layout text must originate from a real paired corpus, we compare three construction strategies on MS‑COCO in Tab.~\ref{tab:text_source_ablation}:
(i) sampling layouts from the real dataset;
(ii) \emph{random} layouts with a minimal bounding‑box size threshold to avoid degenerate cases; and
(iii) layouts generated by GPT‑4o.
Across these settings, reinforcement learning yields \emph{comparable} performance, indicating that Stage\,3 has strong data robustness.

\section{Conclusion}

We present EchoGen, a unified framework for layout‑to‑image generation and image grounding that exploits their duality to achieve accurate layouts and robust grounding in one model. To address naïve joint‑training challenges, we use a progressive scheme: Parallel Multi‑Task Pre‑training for base abilities with shared image‑token pathways, Dual Joint Optimization to serialize generation and grounding into a unified objective that strengthens layout-image-layout loop consistency, and Cycle RL that uses grounding discrepancy as rewards, enabling self‑supervised learning without visual labels.

\bibliography{aaai2026}

\end{document}